\documentclass[lettersize,journal]{IEEEtran}

\usepackage{graphicx}
\usepackage{epsfig}
\usepackage{amsmath}
\usepackage{amssymb}
\usepackage{epstopdf}
\usepackage{amsmath}
\usepackage{makecell}
\usepackage{amsfonts}
\usepackage{color}
\usepackage{multirow}
\usepackage{setspace}
\usepackage{booktabs}
\usepackage{enumerate}
\usepackage{pifont}
\usepackage{ragged2e}
\usepackage{colortbl} 
\usepackage{balance}
\usepackage{bbding}
\definecolor{mygray}{gray}{.95}

\usepackage[linesnumbered,ruled,vlined]{algorithm2e}

\usepackage[colorlinks,urlcolor=black, citecolor=blue]{hyperref}

\newcommand{\blue}{\textcolor[rgb]{0,0.1,0.6}}
\newcommand{\myref}[1]{(\ref{#1})}
\newcommand{\ie}{\emph{i.e.}{}}

\newcommand{\etc}{\emph{etc}{}}

\begin{document}

\title{Synthetic-To-Real Video Person Re-ID}
\author{Xiangqun Zhang, Wei Feng, Ruize Han, Likai Wang, Linqi Song, Junhui Hou
\thanks{
%\revised{This work involved human subjects or animals in its research. Approval of all ethical and experimental procedures and protocols was granted by Ethics Committee of Tianjin University.}
%Likai Wang, Xiangqun Zhang, and Wei Feng are with the College of Intelligence and Computing, Tianjin University, Tianjin 300350, China (The first two authors contribute equally to this work.) (e-mail: kkww@tju.edu.cn; clzxq@tju.edu.cn; wfeng@ieee.org).

Xiangqun Zhang, Wei Feng, Likai Wang are with the School of Computer Science and Technology, College of Intelligence and Computing, Tianjin University, Tianjin, China.

Ruize Han is with the Shenzhen Institute of Advanced Technology, CAS, and the Department of Computer Science, City University of Hong Kong. (Corresponding Author, e-mail: rz.han@siat.ac.cn).

Linqi Song and Junhui Hou are with the Department of Computer Science, City University of Hong Kong, Hong Kong SAR.
}}

\markboth{SUBMITTED}%
{From Synthetic to Real Unveiling the Power of Synthetic Videos for Video Person Re-ID}

\maketitle

\begin{abstract}
	Person re-identification (Re-ID) is an important task and has significant applications for public security and information forensics, which has progressed rapidly with the development of deep learning.
	In this work, we investigate a novel and challenging setting of Re-ID, \ie, cross-domain video-based person Re-ID.  
	Specifically, we utilize synthetic video datasets as the source domain for training and real-world videos for testing, notably reducing the reliance on expensive real data acquisition and annotation. %Technically, to unveil the power of synthetic data for video person Re-ID,
	To harness the potential of synthetic data, we first propose a self-supervised domain-invariant feature learning strategy for both static and dynamic (temporal) features. %Then, to further improve the person identification ability in the target domain, we develop a mean-teacher scheme with a self-supervised ID consistency loss.
	Additionally, to enhance person identification accuracy in the target domain, we propose a mean-teacher scheme incorporating a self-supervised ID consistency loss.
	Experimental results across five real datasets validate the rationale behind cross-synthetic-real domain adaptation and demonstrate the efficacy of our method.
	%verify the rationality of cross-synthetic-real domain adaption and the effectiveness of our method. 
	%The finding that the synthetic data performs even better than the real data in the cross-domain setting is surprising.
	Notably, the discovery that synthetic data outperforms real data in the cross-domain scenario is a surprising outcome. The code and data are publicly available at \url{https://github.com/XiangqunZhang/UDA_Video_ReID.}
	
	% \keywords{complementary view \and multi-human association \and  multi-human tracking \and benchmark}
\end{abstract}

\section{Introduction}
\label{sec:intro}

\begin{figure}[t]
	\centering
	\includegraphics[width=0.95\columnwidth]{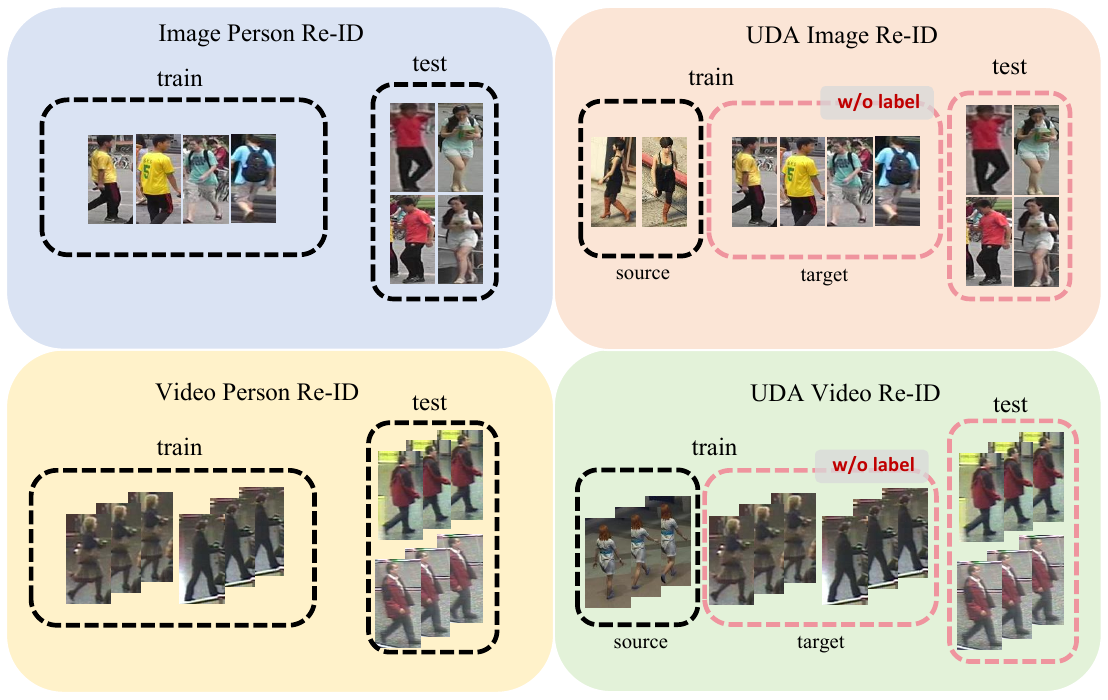} 
	\caption{Illustration of different settings for person Re-ID tasks, including the (single domain) image/video person Re-ID and unsupervised domain adaptive (UDA) image person Re-ID in previous works. In this work, we focus on a new task of video-based UDA person Re-ID. Specifically, we are interested in the scene using synthetic videos as the source domain, which is more economical and practical.}
	\label{fig1}\vspace{-0pt}
\end{figure}

Person re-identification (Re-ID) stands as a classical problem in artificial intelligence, which holds significant real-world applications for public security and information forensics, \textit{e.g.}, intelligent video surveillance and criminal investigation~\cite{he2023region,wang2022occluded}. 
This task experiences rapid evolution in recent years.
We first review the development of person Re-ID as shown in Fig.~\ref{fig1}.
On one hand, in terms of the data format, early works mainly use a single image for retrieving the specific person~\cite{li2014deepreid,ahmed2015improved}.
Later, the sequential video data are applied to person Re-ID since it contains more plentiful information~\cite{you2016top,ouyang2019video,chen2020temporal}, \textit{e.g.}, the human gait.
On the other hand, in terms of the dataset conditions, early works focus on 
%developing the algorithms on the specific datasets by splitting them into training and testing datasets.
%This way, 
the data distributions in the training and testing datasets are similar.
In recent works, unsupervised domain adaptive (UDA) person Re-ID is proposed aiming to train the model on the labeled source domain and unlabeled target domain data, and test it on the target domain. This is more practical since the characteristics of data in the training and testing sets are always various in the real-world application, including the illumination, viewpoint, \textit{etc.}
For UDA person Re-ID, several works~\cite{zhang2021unrealperson,sun2019dissecting} apply the synthetic data for model training to alleviate the burden of data collection and annotation.

Based on the above investigation, we find that the research on cross-domain person Re-ID predominantly focuses on image-based methods, while \textit{the exploration of domain adaptive person Re-ID for videos is limited}, not to mention using synthetic data as the training data.
Naturally, there is an urgent need to leverage richer temporal videos for synthetic data-based cross-domain person Re-ID.   
This way, we aim to explore a question: \textbf{For the video-based person Re-ID problem, can we utilize the synthetic datasets for training and cross-domain (dataset) testing, to address the challenges of annotating and constructing huge video datasets in real-world scenarios? }  
By leveraging synthetic data, we aim to learn robust person representations that are invariant to domain variations, leading to significant improvements in the applicability of video Re-ID.   
However, the synthetic data-based cross-domain video Re-ID brings new challenges.
The first one is \textit{how to learn the domain-invariant features from the synthetic videos}.
Second, learning \textit{a stable and robust video Re-ID model} applied to real data is important yet challenging.

To tackle these challenges, we propose the cross-synthetic-real video person Re-ID framework.
First, specifically for the synthetic source domain, we try to excavate the ability of synthetic data for network training.
To achieve this, we devise a domain-invariant feature learning module that employs an adversarial training strategy to effectively learn the domain-invariant features \textit{from the synthetic source domain to the real target domain}. 
%\JHDEL{, which is obtained through the self-supervised domain classification and gradient reversal layer (GRL).} 
This process is facilitated through self-supervised domain classification and a gradient reversal layer (GRL).
Also, focusing on this video Re-ID task, we particularly consider the \textit{video-level domain invariant features} by combining multiple frames as a sequence, and further multiple sequences as long videos.
By integrating three self-supervised auxiliary tasks, this module {learns the domain-invariant feature for both single frame and whole video}.
Further, we develop the \textit{self-supervised ID consistency guided mean-teacher scheme} to improve the model reliability trained on synthetic data for application to the real data.
This scheme includes a teacher network containing the history information to `supervise' the student network training. 
We specifically design a clustering-based ID consistency estimation method and further develop the 
{self-supervised ID consistency guided mean-teacher} network training.

The proposed method has two typical characteristics for the UDA video Re-ID.
First, our method is specifically designed for the video task, considering the \textit{inter-frame relationship is important for this video ReID task}. 
In the multi-level feature learning strategy, besides the classical image-level feature, we also learn the video-level domain invariant features, to enhance the temporal feature learning in the video task.
%which \textbf{is specifically designed for the video}.
Second, this work specifically designs for the synthetic-to-real generalization.
Our basic idea is to leverage the inherent domain gap between synthetic and real data for learning the domain invariant feature with an adversarial training strategy.
%Specifically, we use , composed of the domain classification and GRL, to learn the domain invariant feature.
This makes our method \textit{appropriate to synthetic source domain and real target domain} with a rationale that they are \textit{divisible with domain difference}, \textit{e.g.}, image style, \textit{etc}. 
%The domain invariant feature is adversarially learned via the domain classification. 
But our method is also general to be applied to real source and target domains since they also have implicit differences. %which is verified by the real-to-real ReID results in the experiments.
We also obtain some insightful observations from the experimental results. For example, we find that the cross-domain video ReID {depends not only on the training data scale but also (more importantly) on its variety}, especially the data style and distribution differences between the training and testing datasets. 
Further, the sample richness in the synthetic data {could narrow the domain gap between the synthetic and real data}, to be {lesser than the gap between two real datasets}.

The main contributions of this work are summarized as follows.
%\revised{If our method is specifically designed for the video. As discussed in Section 3.2, besides the first loss (Eq.7) used for image-level feature, considering the \textit{inter-frame relationship is important for this video ReID task}, we develop the video-level domain invariant features by combining multiple frames as a sequence (Eq.8), and further multiple sequences as long videos (Eq.9). 
	%This enhances the temporal feature learning in our video task, which \textbf{is specifically designed for the video}.
	%The relation of our method and synthetic data. Actually we develop the specific design for synthetic data but the method can also be a generalized UDA model. Our basic idea is to leverage the inherent domain gap between synthetic and real data for learning the domain invariant ID feature. Specifically, in Section 3.2, we use an adversarial training strategy, composed of the domain classification and gradient reversal layer (GRL), to learn the domain invariant feature.
	%Our method is appropriate to synthetic source domain and real target domain with a rationale that they are \textit{divisible with domain difference}, \textit{e.g.}, image style, \textit{etc}. The domain invariant feature is adversarially learned via the domain classification. 
	%Note that, this method is general to be applied to real source and target domains, since they also have implicit differences, which is verified by the real-to-real ReID results in Table 3.
	%}
\begin{itemize}
	%	\item \JHDEL{To the best of our knowledge, this is the first work to study the cross-domain video person Re-ID with synthetic data.}\JHDEL{Based on the existing synthetic and real datasets, we build a benchmark for this problem, including the training/evaluation data splitting and comparative methods.}
	\item This study represents the \textit{pioneering} work in exploring cross-domain \textbf{video} person Re-ID using \textbf{synthetic} data. By leveraging existing synthetic and real datasets, we establish a benchmark for this task, encompassing data partitioning for training and evaluation, as well as comparative methodologies. 
	%	\item \JHDEL{We develop a new method for this problem. The proposed model utilizes the domain difference and relation between the synthetic videos (source domain) and real videos (target domain), to effectively learn the video-level domain-invariant features and real-data adaption, which are achieved through the designed self-supervised losses.} 
	\item	Our novel approach addresses this challenge by capitalizing on the discrepancies between synthetic videos (source domain) and real videos (target domain) to extract\textit{ video-level domain-invariant} features and facilitate real-data adaptation. These objectives are accomplished through self-supervised losses integrated into our model design. 
	% evaluation, both of  .
	\item Extensive experimental results demonstrate the effectiveness of the proposed method, which outperforms the comparative methods over 5.1\%, 11.2\%, 11.3\%, 2.9\%, and 4.1\% mAP score on five real evaluation datasets.
	Further analysis also verifies that the utilization of synthetic videos for cross-domain video person Re-ID is effective, even better than real videos.
\end{itemize}

The remainder of this paper is organized as follows. 
Section~\ref{sec:rela} reviews the related works. 
Section~\ref{sec:method} elaborates on the proposed approach in detail. 
Section~\ref{sec:exp} provides the experimental results and analysis. After that, we provide a brief conclusion including the future work in Section~\ref{sec:con}. 
%\JHNOTE{As a journal paper, it is better to add the organization of the remaining sections.}

%This method holds the potential to overcome the difficulties associated with annotating and constructing datasets in real-world scenarios, providing a solution to the common challenges in video-based pedestrian re-identification tasks.

\section{Related Work}
\label{sec:rela}

\textbf{Video-based person Re-ID} is different from image-based Re-ID, which relies on videos as input and contains richer spatial-temporal information compared to static images. 
Therefore, exploring spatial-temporal cues is crucial in video-based person Re-ID problems. 
With respect to the stable spatial representation, the method in~\cite{gu2020appearance} builds the appearance-preserving module to solve the problem of appearance destruction. 
Another work~\cite{bai2022salient} combines salient-to-broad attention and interaction-driven message passing to capture spatial information. 
Specific to the challenge of changing clothes, CCVID~\cite{gu2022clothes} proposes a clothes-based adversarial loss to mine clothes-irrelevant features for clothes-changing video person Re-ID. 
Similarly, to obtain more effective temporal representations, many technologies are developed, such as complementary features of successive frames~\cite{hou2020temporal}, temporal reciprocal learning for temporal information accumulation~\cite{liu2021watching}, and temporal shift attention module~\cite{wu2022cavit}. 
Furthermore, some approaches focus on aggregating both spatial and temporal features to address spatial and temporal distractors. A spatial-temporal aggregation module is proposed to enhance the aggregation capability~\cite{wang2021pyramid}. Another work~\cite{eom2021video} leverages spatial-temporal memories to refine frame-level person representations to get sequence-level representation. The axial-attention strategy is utilized to capture spatial and temporal relations for addressing the misalignment issue in video-based Re-ID~\cite{liu2021video}. DCCT~\cite{liu2023deeply} develops a deeply-coupled convolution-transformer with spatial-temporal complementary learning for video-based person Re-ID.
These methods aim to fuse features from multiple temporal and spatial scales to obtain comprehensive representations. 
More recently,  the multi-modal pre-trained models are also used to the Re-ID task, such as TF-CLIP~\cite{tfclip}. 
We find that although significant progress has been made in \textit{video person Re-ID, in which cross-domain scenarios have been relatively limited explored}, not to mention that across the synthetic and real domains.

\noindent\textbf{Unsupervised domain adaptive (UDA) image-based person Re-ID} relies on transferring knowledge learned from the source domain to the target domain. First, some approaches enhance the accuracy of representation learning in the target domain by improving clustering methods. The method in~\cite{zheng2021group} utilizes online clustering and label optimization to improve the accuracy of representation learning in the target domain. Another work~\cite{hu2022divide} considers the multi-camera distribution and temporal continuity in re-ID to effectively reduce clustering errors. 
A density-based clustering with adaptive sample augmentation and discriminative feature learning is utilized to estimate and augment person clusters in target domains~\cite{zhai2020ad}. 
Another work~\cite{zheng2021online} dynamically and progressively updates the pseudo labels based on hierarchical cluster dynamics. Liao et al. propose an efficient mini-batch sampling method, called graph sampling (GS), for large-scale deep metric learning~\cite{liao2022graph}.
Second, some methods focus on generating and refining pseudo-labels to mitigate the impact of label noise in the target domain. The authors of \cite{wu2022multi} propose a multi-centroid memory module that captures different ID information within a cluster. Han et al. \cite{han2022delving} propose a probabilistic uncertainty-guided progressive label refinement approach. 
The work in~\cite{li2022reliability} utilizes a mean teacher model for online label refining. Another work \cite{cho2022part} leverages both global and local context information to mitigate label noise.
Third, some methods reduce the domain offset between the source domain and the target domain by domain distribution alignment, \textit{e.g.}, the graph convolutional network-based multi-domain information fusion module~\cite{bai2021unsupervised}, and feature spaces constructing and alignment with the soft labels~\cite{yu2019unsupervised}, the extraction of a support set composed of source images that maximizes the similarity with the target data~\cite{rami2024source}.
The advancements in UDA methods have significantly contributed to cross-domain person Re-ID. 
However, these methods predominantly focus on \textit{image-based tasks and have not yet explored video-based tasks}.   
Differently, this work tackles the challenging task of cross-domain Re-ID in the context of video data, aiming to learn robust and invariant person representations during the temporal variations.

\begin{figure*}[t]
	\centering
	\includegraphics[width=0.95\textwidth]{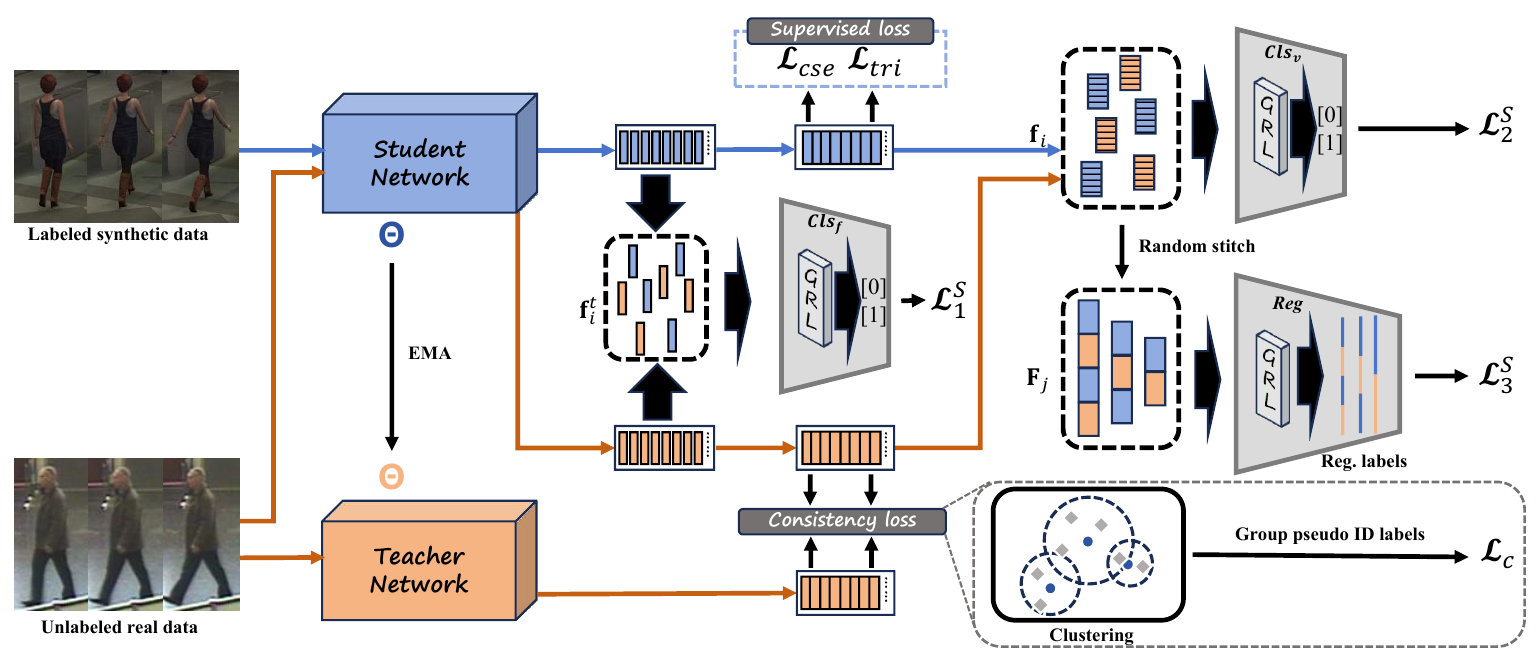} 
	\caption{Framework of the proposed method. We use the synthetic videos with ID labels (source domain) and the real videos (target domain)  without annotation as input. The main network is trained using the supervised losses on the source domain data. Also, for video representation learning on the domain-mixed data, we design a series of self-supervised losses $\mathcal{L}^{\rm S}$ to distinguish the domain labels under different levels, including frame, video, and combined video. We further apply a mean-teacher strategy for the network training on the unlabeled real data. By taking the main network as the student, we use the EMA to obtain a teacher network to ``supervise'' it, which is achieved by the self-supervised consistency loss $\mathcal{L}^{\rm C}$ between them. }
	\label{fig2}
\end{figure*}

\noindent\textbf{Person Re-ID with synthetic data.}
%Some works also use data synthesis and augmentation to increase the diversity of samples for Re-ID. 
In recent years, the difficulty of labeling real data has made synthetic data shine in the field of person Re-ID~\cite{10.1007/s11263-024-02102-x}.
The work in~\cite{bak2018domain} proposes to synthesize realistic images and minimize the domain shift between synthetic and real images.
Later, a number of large-scale image-based synthetic datasets \cite{sun2019dissecting,10.1145/3394171.3413815,zhang2021unrealperson,wang2022cloning} have been proposed for person Re-ID, which provide realistic appearance and clothing by generating virtual human models as a cost-effective alternative to collecting and labeling real data.
More recently, the diffusion based method is also applied to the synthetic Re-ID data generation~ \cite{niu2024synthesizingefficientdatadiffusion}.
A series of methods~\cite{xiang2022rethinking,xiang2020unsupervised,kang2022synthetic} for cross-domain person Re-ID have demonstrated the effectiveness of cross-domain design from synthetic to real.   
However, as discussed above, research on \textit{synthetic data-based person Re-ID mainly focuses on image-based tasks}, \textit{the exploration of which using video data is limited and worth studying}. 
%These methods improve the performance of person Re-ID in different domains by utilizing synthetic data.   
A recent work \cite{wang2022benchmark} presents a large-scale video-based synthetic person Re-ID dataset that demonstrates the potential of synthetic data for person Re-ID in challenging scenarios, which is available for the synthetic-to-real video Re-ID task in this work.

\section{Proposed Method}
\label{sec:method}

%\subsection{Overview}

\subsection{Problem formulation}

We first formulate the unsupervised domain adaptive (UDA) video-based person Re-ID problem.
Following the classical UDA problem, we use the videos in the source domain with ID labels
\begin{equation}
	\mathcal{V}^\mathrm{L}=\{(x_{i}, y_{i})\vert {i=1,...}{,N_L} \},
\end{equation} 
where $N_L$ denotes the number of sequences. For the $i$-th sequence $x_{i}$ in the source domain dataset, we can obtain the corresponding person ID label $y_{i} \in \{1,2,\ \cdots,  C^{ \rm L}\}$, where $C^{ \rm L}$ is the number of whole person IDs. We also use the unlabeled videos in the target domain
\begin{equation} 
	\mathcal{V}^{ \rm U}=\{(u_{j})\vert {j=1,}...{,N_U} \},
\end{equation}
as input, where $N_U$ represents the number of sequences in the target-domain training dataset, but the identity label of each sequence $u_{j}$ is unavailable.
With the above training data, we aim to estimate the ID of each sequence $\mu_{k}$ in the testing dataset
\begin{equation} 
	\mathcal{V}^{ \rm T}=\{(\mu_{k})\vert {k=1},...,{N_T} \}.
\end{equation}

In this work, we specifically study the UDA problem from synthetic data to real data. 
This way, we use the synthetic videos as the source domain $\mathcal{V}^{ \rm L}$, and the real-world videos are used as the unlabeled training set $\mathcal{V}^{ \rm U}$ and testing set $\mathcal{V}^{ \rm T}$.

\subsection{Multi-level domain invariant feature learning}
\label{sec:multi-level}

\noindent\textbf{Basic feature extraction.}
Given the videos $\mathcal{V}^\mathrm{L}$ in the source domain, we first apply an existing deep learning model, \textit{e.g.}, SINet~\cite{bai2022salient}, as the backbone for video Re-ID feature extraction
\begin{equation}
	\label{eq:backbone}
	\mathbf{f}_i = \mathrm{F}(x_i|\Theta), \quad \mathbf{p}_i = \phi (\mathbf{f}_i),
\end{equation}
where $\mathrm{F}(\cdot |\Theta)$ denotes the feature extraction function with the parameters $\Theta$. 
$\mathbf{f}_i$ is the extracted feature of the video sample $x_i$. 
After that, we use the classifier function $\phi (\cdot )$, which is composed of the FC layers with the `\textit{softmax}' activation function, to map the feature into the ID prediction vector $\mathbf{p}_i\in \mathbb{R}^{C^\mathrm{L}}$. 
The probabilities of $x_i$ belonging to all $C^\mathrm{L}$ identities are provided by $\mathbf{p}_i$.

\vspace{1em}
\noindent\textbf{Supervised identification loss.}
Given the person ID label $y_{i}$ in the source domain, we can train the above models $
\mathrm{F}$ and $\Theta$ using the supervised manner. Specifically, we minimize the discrepancy between predicted results and the ground-truth ID. We first denote a one-hot vector $\mathbf{y}_{i}$ with $\mathbf{y}_{i}(c) = 1$ if the index $c$  equals to the correct ID label $y_{i}$, otherwise $\mathbf{y}_{i}(c) = 0$.	
We apply two classical losses, \textit{i.e.}, the cross-entropy loss and the triplet loss as below
\begin{equation}
	\label{eq:Lcse}
	\mathcal{L}_{\rm cse}= - \sum\nolimits_{c} \mathbf{y}_{i}(c) \log \mathbf{p}_i(c),
\end{equation}	
%	$p(y_{i}^\mathrm{s}|x_{i}^\mathrm{s})$ (taken from $\mathbf{p}_i^\mathrm{s}$) denotes the predicted possibility of $x_{i}^\mathrm{s}$ taking the correct label $y_{i}^\mathrm{s}$. 
\begin{equation}
	\label{eq:Ltri}
	\mathcal{L}_{\rm tri} = [\mathrm{D}(\mathbf{f}_a, \mathbf{f}_p) - \mathrm{D}(\mathbf{f}_a, \mathbf{f}_n) + m]_{+} ,
\end{equation}
where $\mathbf{f}_a$, $\mathbf{f}_p$, and $\mathbf{f}_n$ denote the feature of the anchor, positive, and negative sample, respectively. We use $\mathrm{D}(\cdot, \cdot)$ to measure the Euclidean distance between the two feature vectors, and $m$ is the margin used in triplet loss.

\vspace{1em}
\noindent\textbf{Self-supervised domain discrimination loss.}
Motivated by recent advancements in self-supervised learning, alongside the above supervised Re-ID loss, we also introduce a self-supervised domain adaptive loss to manage cross-domain differences, for which we design the self-supervised auxiliary tasks without external annotations.
Specifically, our focus lies on concurrently addressing cross-domain and temporal feature learning, aiming for effective cross-domain video person Re-ID and learning pedestrian features resilient to domain distinctions and temporal variations.
For this purpose, we develop a simple and effective strategy by integrating three self-supervised auxiliary tasks for domain discrimination, as shown in Fig.~\ref{fig2}.

We use both the videos $\mathcal{V}^\mathrm{L}$ in the source domain and the videos $\mathcal{V}^\mathrm{U}$ in the target domain. 
We first get the frame-level feature $\mathbf{f}_i^t$ of sequence  $x_i$ (or $u_j$), which denotes the feature of one frame at time $t$.
Then, we apply a binary domain prediction task to predict the domain of each frame-level feature
\begin{equation}
	\label{eq:frm}
	\mathbf{d}_i^t = \mathrm{Cla}_f(\mathbf{f}_i^t),
\end{equation}
where $\mathbf{d}_i^t$ denotes the domain prediction result of $\mathbf{f}_i^t$, \textit{i.e.}, source domain or target domain. The function $\mathrm{Class}_f$ is a binary classification function.
Similarly, we also consider the video-level domain discrimination as
\begin{equation}
	\label{eq:seq}
	\mathbf{d}_i = \mathrm{Cla}_v(\mathbf{f}_i),
\end{equation}
where $\mathbf{f}_i$ denotes the feature of input sequence, $\mathbf{d}_i$ denotes the binary domain prediction result of video $\mathbf{f}_i$, and $\mathrm{Cla}_v$ also denotes the binary classification. 
{The domain label of each frame or video is easy to obtain; hence the above training can be conducted in a self-supervised manner.}
Besides, we further stitch the features from multiple sequences with a random permutation, as shown in Fig.~\ref{fig2}. 
The number of sequences for combination is arbitrary, \textit{e.g.}, 2,4,8, \textit{etc.}
Then, we aim to predict the domain distribution of the combined features.
The prediction can not be achieved by a classification problem since the number of sequences in each combined feature is various.
This way, we design a regression strategy for the domain prediction.
Specifically, for $V$ sequence features $\mathbf{f}_i$, we use the temporal feature aggregation method~\cite{Chen_2020_CVPR} to form the combined feature as $\mathbf{F}_g$, where $g$ denotes the index. 
Then we predict its domain distribution by
\begin{equation}
	\label{eq:cseq}
	\mathbf{d}^F_g = \mathrm{Reg}(\mathbf{F}_g),
\end{equation}
where $\mathbf{d}^F_g$ is a one-dimensional vector with a fixed length.
Given the domain label of the video sequences, we can also get the domain distribution of the combined videos, as shown in the right of Fig.~\ref{fig2}.

The total self-supervised loss for domain discrimination is defined as
\begin{equation}
	\begin{aligned}
		\label{eq:Ldomain}
		\mathcal{L}^{\rm S} &= \mathcal{L}^{\rm S}_1 + \mathcal{L}^{\rm S}_2 + \mathcal{L}^{\rm S}_3 \\ &= \sum_i\sum_t \mathcal{L}_{\rm cse}(\textbf{d}_i^t,\tilde{\textbf{d}_i^t}) + \sum_i \mathcal{L}_{\rm cse}(\textbf{d}_i,\tilde{\textbf{d}_i})  
		\\ &+ \sum_g \mathcal{L}_{\rm mse}(\textbf{d}_g^F,\tilde{\textbf{d}_g^F}),
	\end{aligned}
\end{equation}
where $\tilde{\textbf{d}_i^t}$, $\tilde{\textbf{d}_i}$ and $\tilde{\textbf{d}_g^F}$ denote the label of ${\textbf{d}_i^t}$, ${\textbf{d}_i}$ and ${\textbf{d}_g^F}$, respectively. Here we use the cross-entropy loss $\mathcal{L}_{\rm cse}$ for classification in $\mathcal{L}^{\rm S}_1$ and $\mathcal{L}^{\rm S}_2$, and the MSE loss $\mathcal{L}_{\rm mse}$ for regression in $\mathcal{L}^{\rm S}_3$.

Note that, during training, we apply the adversarial training strategy by using a gradient reversal layer (GRL)~\cite{ganin2015unsupervised}, which reverses the gradient signs during back-propagation.
The basic insight is that, with GRL, the domain discrimination ability gets stronger as the consistency of cross-domain features increases. 
This way, the classification will be optimized to gradually align the feature distributions between the two domains.
We find that the method is effective in this task since the synthetic and real domains are more discriminative than those with both real source and target domains, which boosts the domain invariant feature learning.
By integrating these three auxiliary tasks, we achieve the \textit{multi-granularity domain-invariant feature learning} from the single frame, single video, and stitched multiple videos.
This enables us to handle the invariant feature learning under cross-domain and over-time variations effectively. 

%At the video level, we employ a video frame sequence domain prediction task to learn the temporal invariant features that are crucial for the main task. By correctly predicting the domain ordering in long videos consisting of shuffled clips from the source and target domains, we ensure the acquisition of domain-invariant representations. 
%One of the primary challenges in video-based cross-domain person re-identification is the temporal and spatial variations in human behavior that lead to domain distribution discrepancies.   For instance, different individuals can perform the same action in completely different ways due to personalized spatio-temporal styles.

%Recently, the progress of adversarial domain adaptation (DA) methods in addressing cross-domain image problems using domain discriminators trained through adversarial learning is noteworthy. Treating domain recognition as a self-supervised auxiliary task is viable since domain labels can be self-annotated. However, directly applying image-based DA to video tasks yields suboptimal results due to the disregard of temporal information. Thus, the question arises: "How can self-supervised auxiliary tasks be designed to facilitate fine-grained cross-domain operations?" More

%In order to overcome the problem of temporal and spatial variations in human behavior and the resulting domain distribution discrepancies in pedestrian re-identification, we propose a comprehensive framework that leverages auxiliary tasks to address cross-domain challenges at different scales. 

\subsection{Consistency learning on unlabeled real data}
\label{sec:CD}
\textbf{{Self-supervised ID consistency aware mean-teacher scheme}.}
As shown in Fig.~\ref{fig2}, we apply the mean-teacher strategy for the learning on unlabeled real data.
Specifically, based on the basic network trained with supervised and self-supervised losses as discussed above, we take it as the `Student' network $\mathcal{S}$ and distill it as the `Mean Teacher' network $\mathcal{T}$
\begin{equation} 
	\mathcal{T} = \mathrm{EMA} (\mathcal{S}),
\end{equation}
where $\mathrm{EMA}$ denotes the exponential moving average (EMA) approach.
In this scheme, teacher $\mathcal{T}$ contains the history information of $\mathcal{S}$,
This way, $\mathcal{T}$ can be utilized to implement backward supervision for guiding the training of student $\mathcal{S}$ on real data without supervised label. We design the consistency loss between $\mathcal{T}$ and $\mathcal{S}$ for self-supervision.

Given the real video sequence $u_j$ in $\mathcal{V}^{ \rm U}$, we first extract the feature of each video
\begin{equation} 
	{\bf{f}}^{\mathcal{S}}_j = \mathcal{S} (u_j), 	\ \ {\bf{f}}^\mathcal{T}_j = \mathcal{T}(u_j),\ \  {j=1,}...{,N_U} ,
\end{equation}
where ${\bf{f}}^{\mathcal{S}}_j$ and ${\bf{f}}^{\mathcal{T}}_j$ denote the feature through the student and teacher networks, respectively.
A straightforward idea to estimate the consistency between $\mathcal{T}$ and $\mathcal{S}$ is to calculate the distance between ${\bf{f}}^{\mathcal{S}}_j$ and ${\bf{f}}^{\mathcal{T}}_j$. 
This may lead to the gradient computation issue because of the large feature distance.
As discussed in previous works~\cite{Chen_2020_CVPR}, a better manner to estimate the consistency is to use the result of the target task rather than the middle features.
This way, a plausible approach is to apply the classification based on the features to obtain the person IDs.
However, this brings two problems.
First, the number of IDs in dataset $\mathcal{V}^{ \rm U}$ is not provided in prior, \textit{i.e.}, the number of classification categories is unknown.
Second, the ID indexes obtained from the classification of $\mathcal{T}$ and $\mathcal{S}$ are unaligned.
To address the above problems, we propose a clustering-based identification consistency estimation strategy.

\vspace{1em}
\noindent\textbf{Clustering-based ID consistency estimation strategy}. We first take all the features ${\bf{f}}^{\mathcal{S}}_j$ and ${\bf{f}}^{\mathcal{T}}_j$ together and apply a clustering algorithm to group these features into $M$ clusters
\begin{equation}
	\label{eq:cluster} 
	\{C_1,C_2,...,C_{M}\} = \mathrm{Cluster}({\bf{f}}^{\mathcal{S}}_j, {\bf{f}}^\mathcal{T}_j), \ \  {j=1,}...{,N_U},
\end{equation}
where $C$ denotes the cluster, which contains different features.
Note that, this process aims to cluster the samples with the same ID into the same group, where the cluster index can be regarded as the pseudo ID label.
With the clustering result, we first define the ID consistency loss as 
\begin{equation} 
	\label{eq:Lc1}
	\mathcal{L}^\mathrm{C}_1 = \sum_{j=1}^{N_U} {\Pi}(C\{{\bf{f}}^{\mathcal{S}}_j\}, {C\{\bf{f}}^\mathcal{T}_j\}),
\end{equation}
where $C\{\cdot\}$ denotes the cluster index of ${\bf{f}}^{\mathcal{S}}_j$ or ${\bf{f}}^{\mathcal{T}}_j$ in Eq.~\myref{eq:cluster}. ${\Pi}(a,b)$ is the indicative function, which results 0 when $ a=b $, and results 1 when $a \neq b$.
This loss constraint that the feature of sample $j$ from $\mathcal{T}$ and $\mathcal{S}$ to be in the same cluster, \textit{i.e.}, with the same pseudo ID.
We also use an ID similarity loss as below
\begin{equation} 
	\label{eq:Lc2}
	\mathcal{L}^\mathrm{C}_2 = \sum_{r=1}^M\sum\nolimits_j (\mathrm{d}({\bf{\bar{f}}}_r, {\bf{f}}^{\mathcal{*}}_j) |{\bf{f}}^{\mathcal{*}}_j \in C_r), \ \ \mathcal{*}={\mathcal{S},\mathcal{T}},
\end{equation}
where ${\bf{\bar{f}}}_r$ denotes the mean feature in cluster $r$, $\mathrm{d}$($\cdot$,$\cdot$) is the distance function. This loss calculates the feature similarities of all features in the same cluster. 
This loss is used to shorten the distances among the features from $\mathcal{S}$ and $\mathcal{T}$ if they have the same pseudo ID.

Compared to directly computing the feature distance, the above strategy incorporates considerations of the overall data distribution and the relationship information among the samples, which can better constrain the identification consistency. The self-supervised identification consistency loss is defined as
\begin{equation}
	\begin{aligned}
		\label{eq:Lcon}
		\mathcal{L}^{\rm C} &= \mathcal{L}^{\rm C}_1 + \mathcal{L}^{\rm C}_2.
	\end{aligned}
\end{equation}

%\torevise{When computing the consistency loss between the teacher and student networks, our initial approach involved directly calculating the Euclidean distance between their feature outputs as the loss function. However, due to the specific architecture of our networks, we encountered the issue of gradient vanishing when the two outputs were very close to each other. Additionally, if the outputs exhibited large variations, it resulted in significant fluctuations in the loss values, leading to unstable training.}

%\torevise{To address these challenges, we adopted a more stable training strategy. Specifically, we employed a mixed clustering method to aggregate the outputs of both networks and generate cluster labels. Subsequently, we utilized the ID consistency loss defined by Eq.~\myref{eq:Lc1} and Eq.~\myref{eq:Lc2}. to measure the consistency of the network outputs. This calculation method offered a more stable gradient signal, ensuring training stability with reduced fluctuations.}

\subsection{Implementation details}

\textbf{Training strategy.} 
The total loss is defined as two supervised and three self-supervised losses as 
\begin{equation} 
	\mathcal{L} = \mathcal{L}_{\rm cse} + \mathcal{L}_{\rm tri}  + \mathcal{L}^{\rm S} +	\mathcal{L}^\mathrm{C}, 
\end{equation}
which are controlled by a learnable weight vector to determine the ratio among them.
%with the weights determined by the exponential function of the parameter vector. 
The student network is first trained with the supervised losses  $\mathcal{L}_{\rm cse}, \mathcal{L}_{\rm tri}$ and the domain discriminative loss $\mathcal{L}^{\rm S}$.
For the training using the mean teacher strategy, the weights of the teacher network are initialized with those of the student network. 
The EMA method is used to update the weights of the teacher network in each training iteration. 
Then, the consistency loss 	$\mathcal{L}^\mathrm{C}_1, \mathcal{L}^\mathrm{C}_2$ can be applied.
Note that, under the standard UDA setting, in the training stage, both synthetic data with ID annotations and and real data (training set) without any annotation are used for training the network.

\textbf{Inference pipeline.} 
Following the classical mean-teacher schema, during inference, only the trained student network is used for testing and evaluation on the real data (testing set).

\textbf{Network details.}  We use the backbone of a video-based person Re-ID approach, \textit{i.e.}, SINet, as the feature extraction model $\mathrm{F}(\cdot |\Theta)$ in Eq.~\myref{eq:backbone}.  
The supervised learning module employs cross-entropy loss and triplet loss for training in Eqs.~\myref{eq:Lcse} and~\myref{eq:Ltri}, with a margin $m$ set to 0.3 for the triplet loss.
In the self-supervised domain discrimination module, both the source domain and target domain videos ($\mathcal{V}^{ \rm L}$ and $\mathcal{V}^{ \rm U}$) are directly fed into the SINet model. For each video composed of multiple frames, frame-level features are extracted and averaged to obtain video-level features. 
%To achieve self-supervised domain adaptation, a domain prediction task with gradient reversal is introduced for both frame-level and video-level features of the source and target domain videos. 
The domain classifier $\mathrm{Cla}(\cdot)$ consists of a fully connected (FC) layer, a ReLU activation function, and a dropout layer. 
Specifically, $\mathrm{Cla}_f(\cdot)$ and $\mathrm{Cla}_v(\cdot)$ have the same structure with unshared parameters.
During the forward propagation, the input data is computed through the domain classifier to produce the final output features. 
In the backward propagation, the gradients are reversed through the adversarial gradient reversal layer (GRL). 

\begin{figure*}[h]
	\centering
	%	\vspace{-20pt}
	\includegraphics[width=0.9\textwidth]{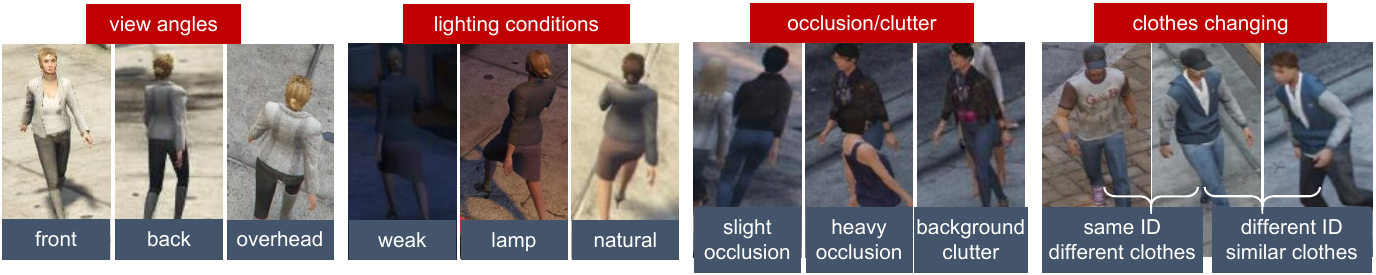} 
	\caption{Illustration of some examples in SVReID of the same identity with various view angles, lighting conditions, occlusion/clutters, and clothes changing. }
	\label{fig4}
	%	\vspace{-25pt}
\end{figure*}

%The video-level features from both networks' outputs are merged for clustering. 
We use the K-means algorithm to partition the video-level features into $M$ clusters, where $M$ is set as 16. 
%The ID consistency loss defined by Eq.~\myref{eq:Lc1} and Eq.~\myref{eq:Lc2} is utilized for training.
We use 4 $\times$ NVIDIA RTX 2080ti for model training and apply the Adam~\cite{kingma2014adam} algorithm for optimization. 
During the testing phase, given a query, the trained student network retrieves a ranked list from the gallery.

\section{Experimental Results}
\label{sec:exp}
\subsection{Setup}

We have constructed {a new benchmark} for cross-synthetic-real video Re-ID task, comprising both synthetic and real datasets. Despite being built upon several existing datasets, this benchmark represents the inaugural initiative for evaluating performance in this specific task.

\textbf{Synthetic video Re-ID dataset.} 
Most previous works like RandPerson~\cite{10.1145/3394171.3413815}, UnrealPerson~\cite{zhang2021unrealperson}, and ClonePerson~\cite{wang2022cloning} for synthetic Re-ID data generation \textit{only focus on the images}.
A recent work \cite{wang2022benchmark} uses the highly realistic video game \textit{Grand Theft Auto (GTA) V} to construct a large-scale dataset for clothes-changing video person Re-ID.
The videos were collected in a surveillance scenario, where 10 monitoring cameras were installed across five different scenes. 
We notice that this dataset mainly focuses on the clothes-changing challenge in the Re-ID problem, in which the persons have various (2-37) suits.
Based on this dataset, we built a \textit{\textbf{S}ynthetic-domain-adaptive \textbf{V}ideo person \textbf{Re-ID}} dataset, namely SVReID. 
Specifically, we first only consider the non-clothing change scenario. 
From the training set in ~\cite{wang2022benchmark}, we extract a subset by selecting the highest number of sequences for each person without clothing variation and get SVReID including 1,164 sequences from 333 identities.
We also consider the clothes-changing problem in some real-world video Re-ID datasets, \textit{e.g.}, CCVID~\cite{gu2022clothes}.
This way, we further extend SVReID by including the clothes-changing videos, \textit{i.e.}, we select the top-8 highest number of sequences with clothing variations and get a total of 3,187 sequences, which form SVReID (w cc).
%Note that, we do note directly use the entire large-scale dataset, since selecting subsets can reduce training time. 
Note that this work mainly focuses on the cross-domain problem, thus we keep the clothing scenario of the training data the same as the corresponding testing dataset. 
This prevents the model from excessively focusing on irrelevant data, and we can compare the performance of different methods under the same scenario.
This synthetic dataset has two main advantages. 
\textbf{Reality}: The videos generated by GTA-V are very \textit{realistic}, as shown in Fig.~\ref{fig4}, the involved persons have various gaits, silhouettes even changed clothes.
\textbf{Diversity}: It offers a wider range of various conditions, such as the camera view angles, illumination, and occlusion/clutter scenarios, which can not be easily achieved in real data.

\vspace{1em}
\noindent\textbf{Real-world video Re-ID datasets.} 
{In this work, we evaluate the effectiveness of our domain-adaptive Re-ID method on four prominent real-world video datasets, \textit{i.e.}, MARS~\cite{zheng2016mars}, iLIDS-VID~\cite{wang2014person}, PRID-2011~\cite{hirzer2011person}, LS-VID~\cite{li2019global}, and CCVID~\cite{gu2022clothes}.\\
	$\bullet$ \textit{MARS} is a large-scale dataset including 1,261 identities and approximately 19,000 video sequences, with an average of 59 frames per sequence. \\
	$\bullet$ \textit{iLIDS-VID} consists of 600 video sequences from 300 unique identities, collected using two cameras, with an average of 70 frames per sequence. \\
	$\bullet$ \textit{PRID}, a smaller dataset, comprises 400 video sequences from 200 identities captured by two cameras, with an average of 108 frames per sequence. \\
	$\bullet$ \textit{LS-VID} is a large-scale dataset including 3772 identities and approximately 14,943 video sequences, with an average of 200 frames per sequence. \\
	$\bullet$ \textit{CCVID} is a Re-ID dataset CCVID derived from a gait dataset FVG~\cite{zhang2019gait}. The characteristic of it is with clothing change setting. CCVID contains 2,856 video sequences from 226 identities, with an average of 2 outfits per person, and an average of 121 frames per sequence. 
}

\vspace{1em}
\noindent\textbf{Evaluation metrics.} In the field of person Re-ID, the evaluation protocols commonly employed are the Cumulative Matching Characteristic (CMC) curve metrics, specifically Rank 1/5/10, and the mean average precision (mAP) score.

\vspace{-0pt}
\begin{table*}[]
	%	\scriptsize
	\caption{Comparison with the state-of-the-art Re-ID methods on four pairs of cross-domain datasets, \textit{i.e.}, from the synthetic SVReID to the real datasets including Mars, iLIDS-VID, PRID, and LS-VID. The values with \textbf{bold} and \underline{underline} mean the best and second best performance, respectively. (\%)} 	\vspace{-0pt}
	\centering \label{tab:com}
	\begin{spacing}{1.05}
		\begin{tabular}{lccccccccc}
			\toprule
			\multirow{2}{*}{Methods}  & \multicolumn{4}{c}{\begin{tabular}[c]{@{}c@{}}SVReID$\rightarrow$Mars\end{tabular}}  & &\multicolumn{4}{c}{{\begin{tabular}[c]{@{}c@{}}SVReID$\rightarrow$iLIDS-VID\end{tabular}}} \\  \cline{2-5} \cline{7-10}
			&  Rank 1 & Rank 5 & Rank 10  & mAP & & Rank 1 & Rank 5 & Rank 10  & mAP  \\ \hline
			CAViT~\cite{wu2022cavit}   & 13.8 & 24.9 & - & 6.5 && 4.0 & 7.3 & - & 6.7 \\ 
			PTSA~\cite{wang2021pyramid} & 34.2 & 50.2 & 56.4 &  20.1 & & 23.7 &42.3 & 52.3 & 30.5 \\
			AP3D~\cite{gu2020appearance} & 44.7 & 61.3 & 66.5 & 25.0 & & 22.7 &39.3 & 50.0 & 31.7\\
			TCLNet~\cite{hou2020temporal} & 42.6 & 57.7 & 64.9 &  22.8 & & 26.7 & 50.0 & 61.3 & 34.9 \\
			% & BiCnet~\cite{hou2021bicnet} & 2.7 & 0.9 & 5.5 & 14.3 & 3.4 & 1.2 & 8.5 & 21.1 \\  
			CAL~\cite{gu2022clothes} & 43.6 & 58.3 & 63.6 & 23.4 & & 25.3 & 41.3 & 51.7 & 31.7 \\
			SINet~\cite{bai2022salient} & 42.8 & 57.6 & 63.1 & 23.4 & & 29.3 & 47.3 & 63.3 & 39.8 \\
			GLT~\cite{zheng2021group} & 38.5 & 55.3 & 62.9 &  24.0 & & 12.0 & 26.0 & 34.0 & 20.0 \\
			QAconv~\cite{liao2022graph} & 40.7 & 55.6 & 62.4 &  22.7 & & 20.0 & 38.0 & 48.0 & 29.4 \\ 
			DCCT~\cite{liu2023deeply} &42.8 & 58.9 & 65.3 &  23.3 & & 24.7 & 45.3 & 61.6 & 32.4 \\ 
			S2P~\cite{rami2024source} & 32.4 & 48.9 & 55.2 &  19.7 & & 18.7 & 35.3 & 43.4 & 27.0 \\ 
			TF-CLIP~\cite{tfclip} &  45.7 & 60.5 & 67.5 &  \underline{25.7} & & 31.3 & 53.7 & 68.7 & \underline{42.1} \\ \hline
			Ours & \textbf{48.1} & \textbf{62.9} & \textbf{68.8} & \textbf{30.8} & & \textbf{42.0} & \textbf{68.0} & \textbf{78.7} & \textbf{53.3}\\
			\toprule
			%\multirow{2}{*}{} & \multirow{2}{*}{Methods} & \multicolumn{4}{c}{CC}
			\multirow{2}{*}{Methods}  & \multicolumn{4}{c}{\begin{tabular}[c]{@{}c@{}}SVReID$\rightarrow$PRID\end{tabular}}  & &\multicolumn{4}{c}{{\begin{tabular}[c]{@{}c@{}}SVReID $\rightarrow$LS-VID\end{tabular}}} \\  \cline{2-5} \cline{7-10}
			&  Rank 1 & Rank 5 & Rank 10  & mAP & & Rank 1 & Rank 5 & Rank 10  & mAP  \\ \hline
			CAViT~\cite{wu2022cavit}   & 13.5 & 38.2 & - & 25.2 && 12.2 & 20.5 & - & 4.1 \\ 
			PTSA~\cite{wang2021pyramid} & 47.9 & 71.4 & 80.1&  51.8  & & 10.6 & 18.1 & 22.5 & 3.7 \\
			AP3D~\cite{gu2020appearance} & 56.2 & 80.9 & 92.1 &  67.4 & & 13.4 & 21.7 & 26.1 & 4.5 \\ 
			TCLNet~\cite{hou2020temporal} & 65.6 & 74.5 & 77.9 &  57.5 & & 11.0 & 19.2 & 23.6 & 3.9 \\
			% & BiCnet~\cite{hou2021bicnet} & 2.7 & 0.9 & 5.5 & 14.3 & 3.4 & 1.2 & 8.5 & 21.1 \\  
			CAL~\cite{gu2022clothes} & 51.6 & 73.0 & 78.7 & 55.1 & & 11.4 & 20.3 & 25.2 & 4.0 \\
			SINet~\cite{bai2022salient} & 53.7 & 75.3 & 83.2 & 63.0 & & 13.8 & 21.7 & 26.4 & 4.7 \\
			GLT~\cite{zheng2021group} & 33.7  & 62.9  & 71.9  &  48.0  & & 9.1 & 16.7 & 20.9 & 3.1 \\
			QAconv~\cite{liao2022graph} & 50.6 & 77.5 & 84.3 &  61.8 & & 11.3 & 20.6 & 25.6 & 4.1 \\
			DCCT~\cite{liu2023deeply} & 49.7 & 74.7 & 81.4 &  59.5 & & 14.3 & 22.5 & 28.3 & 5.3 \\ 
			S2P~\cite{rami2024source} & 40.4 & 67.5 & 84.3 &  53.5 & & 10.1 & 18.1 & 22.3 & 3.4 \\ 
			TF-CLIP~\cite{tfclip} & 63.4 & 72.4 & 88.7 & \underline{67.8} & & 16.2 & 25.6 & 31.8 & \underline{5.7} \\ \hline
			Ours & \textbf{71.9} & \textbf{87.6} & \textbf{89.9} & \textbf{79.1} & & \textbf{20.1} & \textbf{31.9} & \textbf{38.0} & \textbf{8.6} \\
			\bottomrule
		\end{tabular} 
		\label{tab:compar1}
		\vspace{-0pt}
	\end{spacing}
\end{table*}

\begin{table*}[]
	%	\scriptsize
	\caption{Comparison with the state-of-the-art Re-ID methods from the synthetic SVReID with clothes changing, \ie, SVReID (w cc), to the real dataset CCVID.  (\%)} 	\vspace{-0pt}
	\centering \label{tab:com2}
	\begin{spacing}{1.05}
		\begin{tabular}{lccccccccc}
			\toprule
			\multirow{2}{*}{Methods}  & &\multicolumn{4}{c}{{\begin{tabular}[c]{@{}c@{}}SVReID (w cc)$\rightarrow$CCVID\end{tabular}}} \\  \cline{2-5} 
			& & Rank 1 & Rank 5 & Rank 10  & mAP  \\ \hline
			CAViT~\cite{wu2022cavit}  && 61.4 & 70.0 & - & 42.5 \\ 
			PTSA~\cite{wang2021pyramid} & & 61.4  & 74.3 & 80.0 & 52.6 \\
			TCLNet~\cite{hou2020temporal} & & 42.8 & 57.6 & 63.1 & 23.4 \\
			% & BiCnet~\cite{hou2021bicnet} & 2.7 & 0.9 & 5.5 & 14.3 & 3.4 & 1.2 & 8.5 & 21.1 \\  
			CAL~\cite{gu2022clothes}& & 70.7 & 78.1 & 81.6 & 66.8 \\
			SINet~\cite{bai2022salient} & & 70.7 & 79.1 & 83.1 & 64.1 \\
			GLT~\cite{zheng2021group}  & & 68.2  & 77.6  & 80.9  & 59.1 \\
			QAconv~\cite{liao2022graph} & & 55.4 & 70.4 & 75.3 & 37.8 \\
			AP3D~\cite{gu2020appearance}& & 52.8 & 62.6 & 68.2 & 43.1\\ 
			DCCT~\cite{liu2023deeply} & & 66.7 & 78.9 & 80.3 &  57.8 \\ 
			S2P~\cite{rami2024source} & & 53.4 & 65.7 & 72.3 &  41.3  \\ 
			TF-CLIP~\cite{tfclip} & & 72.2 & 81.3 & 84.5 & \underline{68.3} \\ \hline
			Ours & & \textbf{78.3} & \textbf{84.3} & \textbf{87.3} & \textbf{72.4}\\
			\bottomrule
		\end{tabular} 
		\label{tab:compar}
		\vspace{-0pt}
	\end{spacing}
\end{table*}

\begin{table*}[]
	\vspace{-0pt}
	\caption{Ablation study of the proposed method. (\%)} 	\vspace{-0pt}
	\centering
	\begin{spacing}{1.05}
		%	\scriptsize
		\begin{tabular}{lccccccccc}
			\toprule
			\multirow{2}{*}{Methods}  & \multicolumn{4}{c}{\begin{tabular}[c]{@{}c@{}}SVReID$\rightarrow$iLIDS-VID\end{tabular}}  & &\multicolumn{4}{c}{{\begin{tabular}[c]{@{}c@{}}SVReID (w cc)$\rightarrow$CCVID\end{tabular}}} \\  \cline{2-5} \cline{7-10}
			&  Rank 1 & Rank 5 & Rank 10  & mAP & & Rank 1 & Rank 5 & Rank 10  & mAP  \\ \hline
			Baseline   & 29.3 & 47.3 & 63.3 & 39.8 &  & 70.7 & 79.1 & 83.1 & 64.1 \\ 
			+ $\mathcal{L}^\mathrm{S}_1$ loss & 32.7 & 59.3 & 73.3 &  45.4 & & 71.7 & 78.3 & 81.8 & 68.1 \\
			+$\mathcal{L}^\mathrm{S}_1$ + $\mathcal{L}^\mathrm{S}_2$ losses & 34.0 & 62.7 & 75.3 &  47.3 & & 72.9 & 80.5 & 83.1 & 68.2 \\
			+ $\mathcal{L}^\mathrm{S}_1$ + $\mathcal{L}^\mathrm{S}_2$ + $\mathcal{L}^\mathrm{S}_3$ losses & 36.0 & 63.3 & 75.3 &  49.1 & & 73.9 & 79.9 & 84.2 & 70.1 \\
			w/o $\mathcal{L}^\mathrm{C}_1$  loss & 39.0 & 62.7 & 73.3 &  50.1 & & 75.2 & 82.4 & 86.8 & 71.8\\
			w/o $\mathcal{L}^\mathrm{C}_2$ loss  & 39.3 & 64.0 & 78.0 &  51.2 & & 74.6 & 80.3 & 83.8 & 70.0\\
			%		+0-1 loss+group-dis-loss+sf & 0.7 & 4.7 & 7.3 &  4.5 & & 42.3 & 61.3 & 69.1 & 25.6 \\	
			Ours & \textbf{42.0} & \textbf{68.0} & \textbf{78.7} & \textbf{53.3} & & \textbf{78.3} & \textbf{84.3} & \textbf{87.3} & \textbf{72.4}\\
			\bottomrule
		\end{tabular} 
		\label{tab:abl}
		\vspace{-10pt}
	\end{spacing}
\end{table*}

\subsection{Comparison with state-of-the-art methods}

\noindent\textbf{Comparison methods.} {Due to limited research on cross-domain person Re-ID based on videos, we conducted a comprehensive evaluation by comparing our method against two categories of approaches. 
	Similar with the division in the related work section, the first category is the state-of-the-art \textit{video-based person Re-ID approaches}, including CAViT~\cite{wu2022cavit}, PSTA~\cite{wang2021pyramid}, AP3D~\cite{gu2020appearance}, TCLNet~\cite{hou2020temporal},  SINet~\cite{bai2022salient}, CAL~\cite{gu2022clothes},
	DCCT~\cite{liu2023deeply} and TF-CLIP~\cite{tfclip}.  
	%	Among these, CAViT utilizes vision transformers, while CAL focuses on addressing the challenge of clothing variations. 
	The second category is \textit{image-based person Re-ID methods} that aim to handle the \textit{unsupervised domain adaptation (UDA) problem}, including  GLT~\cite{zheng2021group}, QAconv~\cite{liao2022graph} and S2P~\cite{rami2024source}.}

As shown in Table~\ref{tab:com}, we compare our method with the state-of-the-art video person Re-ID approaches and domain adaptive Re-ID approaches. We conduct the experiments on four pairs of cross-domain datasets.
We can first see that the proposed method gets the best performance compared to other approaches with a large margin, on all four real-world datasets.
Specifically, we find that our method outperforms the second-best approach (TF-CLIP) by over 10\% in terms of mAP on iLIDS-VID and PRID datasets.
The performance improvement on the large-scale Mars dataset also achieves over 5\%.
We can also see that on LS-VID the results of all methods are generally poor. 
This is because LS-VID is a large-scale dataset with more challenging testing cases.

As shown in Table~\ref{tab:com2}, on the CCVID dataset, the clothes-changing setting increases the difficulty for Re-ID. 
But since the data variety of CCVID (like the view angles, and illumination) is relatively simple, the overall performance on it is generally high.
In this case, the proposed method still performs better than CAL, which is specifically designed for CCVID to handle the clothes-changing problem.
These results demonstrate that the proposed method has an effective domain adaption ability across synthetic and real-world datasets. 

\subsection{Ablation study}

To verify the effectiveness of each component in the proposed method, we conduct the ablation study on two datasets, \textit{i.e.}, the medium-scale dataset iLIDS-VID and the clothes-changing aware dataset CCVID, as shown in Table~\ref{tab:abl}.

\textbf{Effectiveness of domain-invariant feature learning module} (Section~\ref{sec:multi-level}). We add the self-supervised domain discrimination losses, \textit{i.e.}, $\mathcal{L}^\mathrm{S}_1$, $\mathcal{L}^\mathrm{S}_2$, and $\mathcal{L}^\mathrm{S}_3$ in Eq.~\myref{eq:Ldomain} one by one. As shown in Table~\ref{tab:abl}, we can first see that the proposed self-supervised loss $\mathcal{L}^\mathrm{S}$ is effective in improving the Re-ID performance with a large margin.  Also, the results demonstrate that each part in $\mathcal{L}^\mathrm{S}$ can improve the performance, respectively.  

\begin{table*}[]
	\caption{Comparison results with the real-to-real data training. (\%)}\label{tab:ana}	\vspace{-0pt}
	\centering
	%	\scriptsize
	\begin{spacing}{1.05}
		\begin{tabular}{lcccccccccc}
			\toprule %	\Xhline{1pt} 
			\multirow{2}{50pt}{Methods}  & \multicolumn{4}{c}{Baseline}  & &\multicolumn{4}{c}{Ours} \\  \cline{2-5} \cline{7-10}
			& Rank 1 & Rank 5 & Rank 10  & mAP & & Rank 1 & Rank 5 & Rank 10  & mAP  \\ \hline \hline
			Mars$\rightarrow$iLIDS-VID &6.7 & 12.0 & 15.3 &  \blue{10.7} & & 8.3 & 15.7 & 20.3 & \blue{14.6} \blue {\scriptsize (+3.9)} \\
			{PRID$\rightarrow$iLIDS-VID}  & 16.7 & 29.3 & 41.3 &  \blue{24.4} &  & 24.0 & 42.7 & 54.0 & {\blue{33.5}} \blue {\scriptsize (+9.1)} \\ 
			LS-VID $\rightarrow$iLIDS-VID  & 34.7 & 58.7 & 69.3 & \underline{\blue{46.5}} & & 41.3	&66.5	&76.7	& \underline{\blue{51.4}}  \blue{{\scriptsize (+4.9)}} \\ \hline
			\multirow{2}{60pt}{SVReID \\ $\rightarrow$iLIDS-VID} & 29.3 & 47.3 & 63.3 & \underline{\blue{39.8}} &  & 42.0 & 68.0 & 78.7 & \underline{\blue{53.3}}  \blue{\scriptsize (+13.5)} 	\vspace{-2pt}\\
			&  &  &  &   \underline{\scriptsize ($\downarrow 6.7$)} &  &  &  &  & \underline{\scriptsize ($\uparrow 1.9$)} \\
			\bottomrule %\Xhline{1pt} 
			{Mars$\rightarrow$CCVID}   & 62.6 & 75.3 & 83.3 & \underline{\blue{55.0}} &  & 70.3 & 79.1 & 84.4 & \underline{\blue{60.5}}  \blue{\scriptsize (+5.5)}  \\ 
			PRID$\rightarrow$CCVID  & 43.6 & 58.9 & 64.6 & \blue{33.9} &  & 65.5 & 80.5 & 84.4 & \blue{54.2}  \blue{\scriptsize (+20.3)} \\ 
			LS-VID $\rightarrow$CCVID  & 59.4 & 67.6 & 71.1 & \blue{50.8} & & 66.7	& 75.1	& 80.1 & \blue{60.4} \blue{\scriptsize (+9.6)} \\ \hline
			\multirow{2}{70pt}{{SVReID (w cc) $\rightarrow$ CCVID}}  & 70.7 & 79.1 & 83.1 & \underline{\blue{64.1}} &  & 78.3 & 84.3 & 87.3 & \underline{\blue{72.4}}  \blue{\scriptsize (+8.3)} 	\vspace{-2pt}\\
			&  &  &  &   \underline{\scriptsize ($\uparrow 9.1$)} &  &  &  &  & \underline{\scriptsize ($\uparrow 11.9$)} \\  
			\bottomrule %\Xhline{1pt} 
		\end{tabular}
	\end{spacing}
	\label{tab:analysis}
	\vspace{-0pt}
\end{table*}

\textbf{Effectiveness of cross-domain consistency learning module} (Section~\ref{sec:CD}).
With $\mathcal{L}^\mathrm{S}$ equipped, we further investigate the {effectiveness of the mean-teacher scheme.} 
We can first see that the integration of the mean-teacher training strategy can further improve the performance of our method. 
Also, we remove each corresponding loss component \textit{i.e.}, $\mathcal{L}^\mathrm{C}_1$ in Eq~\myref{eq:Lc1} and $\mathcal{L}^\mathrm{C}_2$ in Eq.~\myref{eq:Lc2} in the self-supervised ID consistency learning, respectively. We can also see that each sub-loss in $\mathcal{L}^\mathrm{C}$ is verified to be useful.

The above ablative results verify the effectiveness of each module in the proposed framework.
Next, we further analyze the rationality of using the synthetic data for cross-domain video person Re-ID by comparing it with real-to-real data training.

\subsection{Cross-dataset evaluation results}
In this section, we also use two real datasets, \textit{i.e.}, iLIDS-VID and CCVID for testing, and use the other three real datasets, \textit{i.e.}, Mars, PRID, and LS-VID for training.
As shown in Table~\ref{tab:analysis}, we compare the cross-domain Re-ID results using the real data (as discussed above) and our synthetic data (SVReID) for training. The experiments are conducted with the baseline and the proposed method, respectively.

\textbf{Impact of the training datasets.}
{First, we compare the results using different real datasets (Mars, PRID, and LS-VID) for training.
	Specifically, on testing of iLIDS-VID, the largest LS-VID obtains the best performance.
	But we find that the {performance of training on MARS is lower} than that on PRID, although the {videos and identifies in PRID are much less than} the videos in MARS.
	However, the results on CCVID are significantly different from those on iLIDS-VID.}
We then analyze the cause of the above results.
iLIDS-VID has various lighting and view-angle variations, which makes both MARS and PRID perform poorly.
MARS shows worse generalization since \textit{its distribution is obviously different} from that of iLIDS-VID, which has also been discussed in MARS paper~\cite{zheng2016mars}.
The LS-VID with more various scenes and identities makes the model perform better.
Different from iLIDS-VID, CCVID is built upon a gait recognition dataset captured by horizontal cameras with similar view angles, which is similar to MARS.
Thus on CCVID, the training on MARS performs better than other datasets.
The above results demonstrate that the performance for cross-domain ReID \textbf{depends not only on the training data scale but also (more importantly) on its variety}, especially the data style and distribution difference between the training and testing datasets.
Even though Mars and LS-VID are large-scale, they can not cover the various scenarios during testing.

Next, we compare the results of using the proposed synthetic SVReID for training and using the real datasets, \textit{i.e.}, Mars, PRID, and LS-VID. 
On iLIDS-VID, for the baseline method, we can see that using SVReID provides lower performance than LS-VID, which indeed has abundant samples. But with respect to the proposed method (`Ours'), SVReID provides a higher performance than LS-VID. This demonstrates the combination of the synthetic dataset and the specific method can greatly improve the performance.
On CCVID datasets, we can see a more significant effect. No matter the baseline or our method, the using of SVReID obtains the best Re-ID performance compared with the other three real datasets. 
Take the mAP score for example, the margin of improvement achieves 9.1\% and 11.9\%, for the baseline approach and our method, respectively.
Such improvements benefit from the synthetic SVReID under various urban scenes with richer samples. 
This demonstrates that the proposed synthetic SVReID has good generalization compared to existing real datasets, which effectively verifies that \textit{the utilization of synthetic videos for cross-domain video person is rational}.

%on which the method also shows consistently high performance on all five testing datasets (as seen in Table~\ref{tab:compar1}).
%This
%Further, by observing the videos in MARS, we find that although the scale is large, the samples are all captured in a campus by similar horizontal view angles and natural lighting, which may have a gap to other datasets, like the indoor iLIDS-VID.

\textbf{Impact of the proposed method.}
We further compare the results of each row in Table~\ref{tab:analysis}.
We can see that our method outperforms the baseline significantly under each setting of training and testing datasets.
This clearly indicates that \textit{the proposed method can improve the cross-domain robustness not only for the scenes from synthetic to real domain but also those from real to real datasets}.

%\subsection{In-depth analysis}
%	\vspace{-20pt}

\begin{table*}[htbp!]
	% \tabcolsep=0.05cm
	%	\tabcolsep=0.07cm
	%	\vspace{-10pt}
	\centering
%	\footnotesize
	\begin{spacing}{1.1}
		\caption{Experimental results of our method on the SVReID and the enlarged SVReID+ datasets. (\%)}
		\label{tab:large}
		\begin{center}
			\begin{tabular}{lcccc|lcccc}
				\hline
				Ours	& R1  &R5  &R10   &mAP &	& R1  &R5  &R10   &mAP \\ \hline
				% CSRD-II & 3K     & 51K & 2 & 5-25        & \ldots \\ \hline\\
				SVReID $\rightarrow$   iLIDS   & 42.0 & 68.0 & 78.7 & 53.3 & SVReID $\rightarrow$ PRID & 71.9 & 87.6 & 89.9 & 79.1  \\
				SVReID+ $\rightarrow$ iLIDS  & 43.3 & 66.7 & 80.7 &55.0  & 	SVReID+ $\rightarrow$  PRID  & 75.3 & 88.8 & 94.4 & 81.0  \\
				\hline
			\end{tabular}
		\end{center}
	\end{spacing}
	%	\vspace{-20pt}
\end{table*}

\begin{figure*}[h]
	\centering
	\includegraphics[width=1\textwidth]{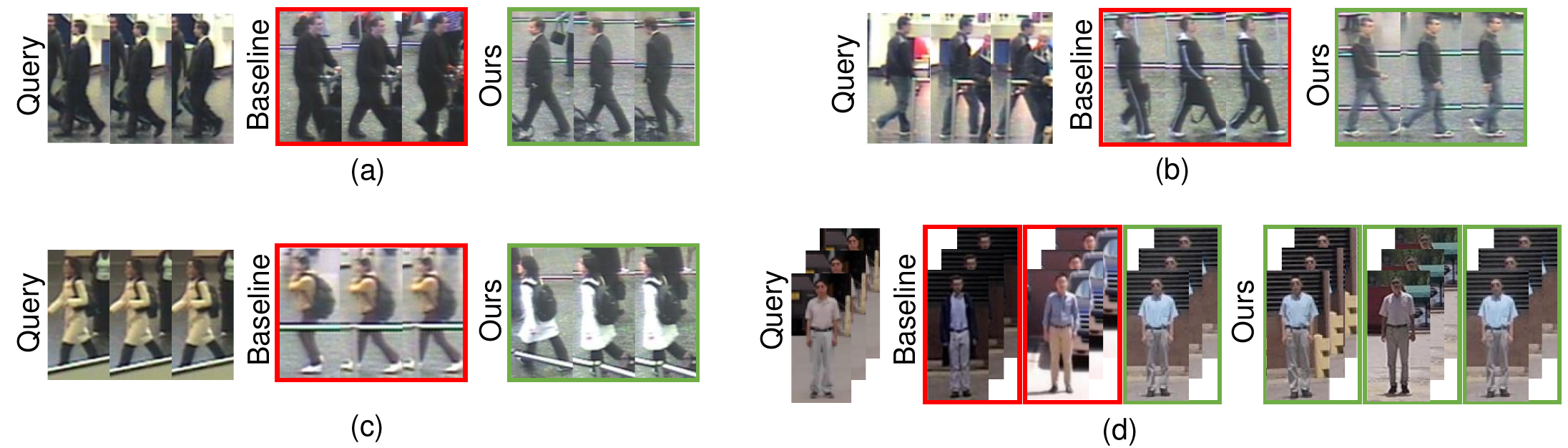} 	\vspace{-15pt}
	\caption{Qualitative analysis of the baseline and the proposed method, under complex background (a), occlusion (b), illumination variation (c), and clothes changing (d).}
	\label{fig3} %\vspace{-10pt}
\end{figure*}

\textbf{Discussion and analysis.} 
We provide an in-depth analysis using synthetic data and our method provides better results than real data.
The first reason is that the proposed SVReID dataset has \textit{abundant samples with various conditions}.
The real datasets PRID and Mars are limited to outdoor footage, iLIDS-VID originates from indoor surveillance, and CCVID, derived from the gait dataset FVG, contains only one view and follows consistent routes.
Their videos are monotonous and independent, which can not cover complex real-world situations.
LS-VID contains relatively various samples. However, the collection and annotation of such a big dataset is high-cost.
By contrast, as shown in Fig.~\ref{fig4}, SVReID has various and controlled scenarios, which offers a wider range of various conditions (gait, silhouette, perspective, illumination, indoor/outdoor scenes, \etc). 
The sample richness \textbf{narrows the domain gap between the synthetic and real data}, which is \textbf{lesser than the gap between two different real datasets}.
Taking the results of SVReID $\rightarrow$ CCVID for example, the training on synthetic SVReID improves the ability to learn Re-ID features irrelevant to clothing changes, \textit{e.g.}, gait, silhouettes, which are important to the clothes-changing dataset CCVID.
Moreover, the construction of SVReID is more economical than the real datasets.
The second reason is that we propose \textbf{the specific solution for synthetic data, which effectively utilizes the domain difference and relation between the synthetic videos (source domain) and real videos (target domain)}, to learn the video-level domain-invariant features and their adaption to the real data.

%\begin{figure}[h]
%	\centering
%	\includegraphics[width=0.7\columnwidth]{figs/fig4.pdf} 
%	\caption{Illustration of some examples in SVReID of the same identity with various styles (a) and clothes changing (b). }
%	\label{fig4}
%	\vspace{-10pt}
%\end{figure}

\subsection{Enlarged SVReID dataset}
{We then consider the data scale of the proposed SVReID.
	We can see that compared with some previous large datasets, such as Mars, and LS-VID, the scale of SVReID is not very large.
	As discussed before, SVReID has a promising generalization ability compared with other real video Re-ID datasets.
	Even so, due to the low cost of synthetic data generation, we further enlarge the proposed dataset to unleash the potential of synthetic data, in this section. 
	Specifically, we enlarge it from 333 identities into 612, and from 1,164 videos into 2,823, to form SVReID+. 
	As shown in Table~\ref{tab:large}, with training on SVReID+, the performance still improves as the data scale expands.
	However, we can see that, with the identities and videos enlarging approximately/over two times, the time of training will rise a lot, but the performance increase is not very significant.
	On the whole, considering the training time and the storage of training data, the SVReID is more economical and promising for the cross-domain video person Re-ID task.
}

\subsection{Qualitative analysis}
Finally, we provide some qualitative results. As shown in Fig.~\ref{fig3}, we illustrate the query and the Rank-1 identities of our method and the baseline approach.
From the case in Fig.~\ref{fig3}(a) we can see that our method can effectively handle the complex background around the person. 
Similarly, from the case in Fig.~\ref{fig3}(b) we can also see that our method can handle the occlusions appearing in the query video. 
The third case demonstrates that our method can deal with the problem of style difference between the query and gallery.
Specifically, we find that the baseline method heavily depends on appearance similarity for identifying individuals, which becomes ineffective when the videos have a large style gap, as shown in Fig.~\ref{fig3}(c). 
Our method addresses this problem since it learns more domain-invariant temporal features from the target itself, which is discriminative and little affected by the surroundings or styles.

We also provide a sample with clothes changing from the CCVID  in Fig.~\ref{fig3}(d).
We find that our model trained on SVReID (w cc) can handle the challenging cloth-changing scenarios.
Figure~\ref{fig4} (last group of samples) shows some examples in the SVReID (w cc) dataset.
Note that, the cloth-changing pedestrian videos are easy to collect in the synthetic environment but difficult in the real world.
Therefore, the usage of synthetic data for the cloth-changing video person Re-ID is very promising.

%\vspace{-12pt}
%Advantage of the synthetic videos for Re-ID.
%\subsubsection{Failure case analysis.}

\section{Conclusion}
\label{sec:con}
In this work, we have studied a new and valuable problem of cross-domain video person Re-ID with synthetic data. For this problem, we developed a new framework, including a domain-invariant feature learning module and a mean-teacher network training scheme, both of which are achieved in a self-supervised manner without extra annotations.
We have also built the benchmark including the training/evaluation datasets and comparative results, for this problem.

A limitation of this work seems that the used training and testing datasets for cloth-changing video person Re-ID is not large and various enough, although the cloth-changing setting is \textit{not} the main purpose of this work. In the future, we plan to build a \textit{unified and large} synthetic dataset, including the clothes unaltering/changing, and other scenarios, for more general video person Re-ID training, for which we also consider using the recently popular diffusion models.   \\

%\textbf{Data availability statements.} 
%The code and data in this work is publicly available at \url{https://github.com/XiangqunZhang/UDA_Video_ReID.}

%\section*{Acknowledgment}
This work was supported by the National Natural Science Foundation of China (NSFC) under Grant 62072334, 62402490 and the Shenzhen Basic Research Foundation under Grant JCYJ20240813154920027.

\bibliographystyle{IEEEtran}
\bibliography{ref}

\end{document}